\documentclass[11pt]{article}

\usepackage[final]{acl}

\usepackage{times}
\usepackage{latexsym}

\usepackage[T1]{fontenc}

\usepackage[utf8]{inputenc}

\usepackage{microtype}

\usepackage{inconsolata}

\usepackage{graphicx}

\usepackage{amsmath}
\usepackage{booktabs}
\usepackage{multirow}
\usepackage{amssymb}
\usepackage{tabularx}
\usepackage[table]{xcolor}
\usepackage{array}

\usepackage{makecell}

\title{Decoupling Generation and Selection for Budget-Constrained Faithful Summarization}

\author{
\textbf{Zeyu Wang} ,
\textbf{Guanghua Wang}\thanks{Corresponding author.}, 
\textbf{Meng Xu}
\\
\ Kean University, USA
\\
\small{
\texttt{\{wangzeyu, guanghua.wang, meng.xu\}@kean.edu}
}
}

\begin{document}
\maketitle
\begin{abstract}

Abstractive summarization models remain vulnerable to factual inconsistency,
redundancy, and weak length control. We propose a modular
generation-and-selection framework for sentence-budget-constrained
summarization. A pretrained generator produces multiple candidate summaries,
which are decomposed into sentence-level candidates. A combinatorial selector
then constructs the final summary by balancing relevance, factuality, and
redundancy under an explicit budget. The framework supports MMR, ILP, and a
DPP-inspired log-determinant objective without retraining the generator.
Experiments on CNN/DailyMail, Multi-News, FaithBench, and TofuEval show
consistent improvements in factuality and source-grounding metrics, especially
for multi-document summarization, at the cost of lower reference-overlap
scores. Human evaluation further indicates higher perceived consistency,
relevance, clarity, and conciseness, with a small reduction in coherence.
These results show that decoupling generation from selection provides a
model-agnostic mechanism for improving factual grounding. Code is available
at \url{https://anonymous.4open.science/r/bcfs-D05E/}.

\end{abstract}

\begin{figure}[t!]
\centering
\includegraphics[width=0.95\columnwidth]{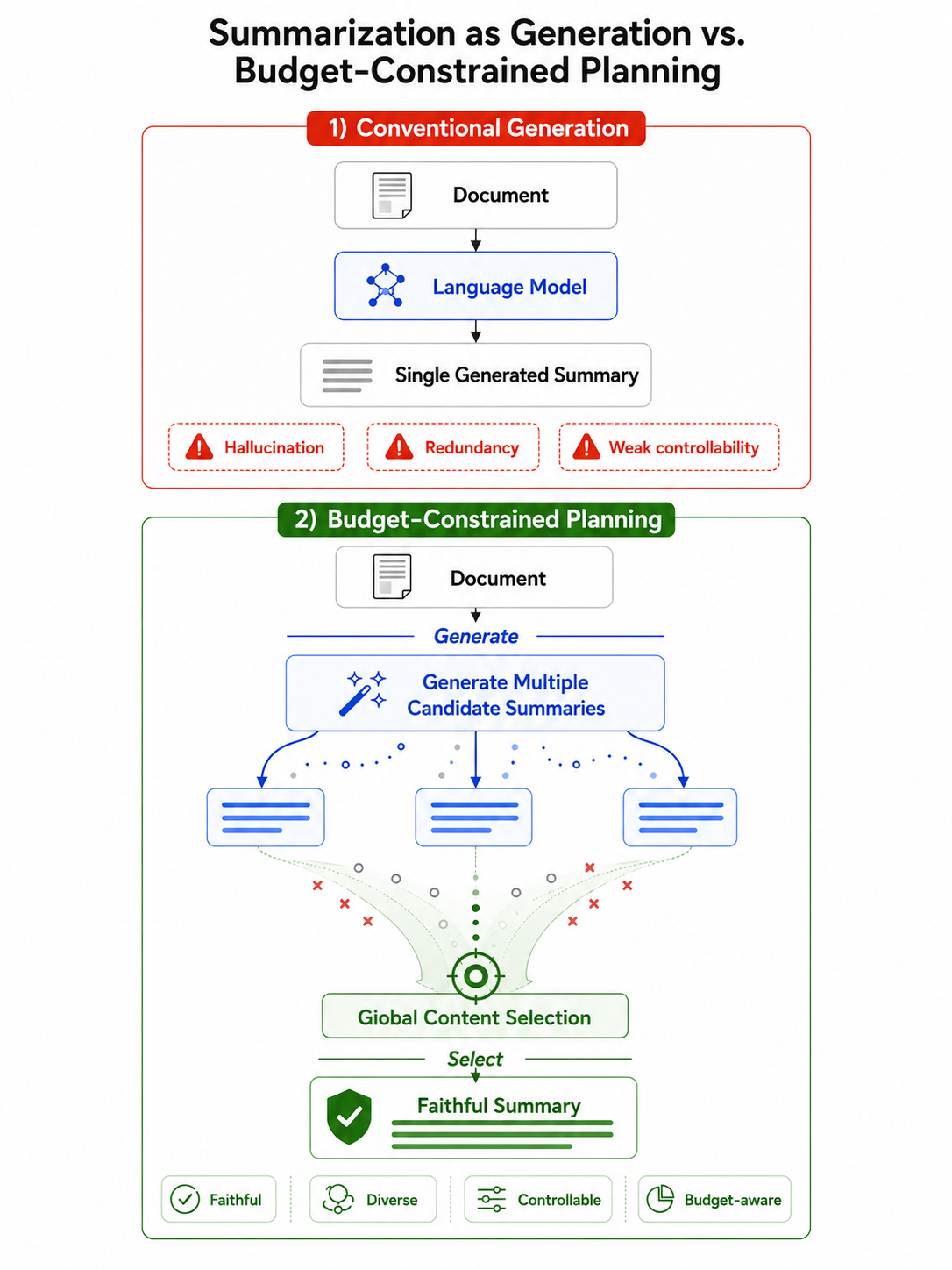}
\caption{
From generation to planning for abstractive summarization.
Conventional autoregressive summarization produces a summary from a single decoding trajectory.
Our framework decouples generation from selection by generating multiple candidate summaries, decomposing them into sentence-level candidates, and selecting a budget-constrained subset that balances coverage, factuality, and redundancy.
}
\label{fig:intro_framework}
\end{figure}

\section{Introduction}

Pretrained language models and large language models (LLMs) have
substantially improved the fluency and semantic quality of abstractive
summarization systems
\cite{li-etal-2024-improving-faithfulness,kim2024fables}.
However, autoregressive generation remains vulnerable to factual
inconsistency, redundant content, and weak control over summary length.
Because content decisions are made along a single token-level decoding
trajectory, generated summaries may contain unsupported claims, omit salient
information, or fail to satisfy a specified output budget
\cite{kim2024fables,tang-etal-2024-tofueval}.

We formulate abstractive summarization as a
\emph{generation-and-selection} problem. Rather than treating a single
generated sequence as the final output, a pretrained generator produces
multiple candidate summaries, which are decomposed into sentence-level
candidates. A combinatorial selector then constructs the final summary under
an explicit sentence budget by balancing source relevance, factual
consistency, and redundancy. The selected sentences are ordered according to
their aligned source positions, requiring neither retraining of the generator
nor an additional rewriting stage.

The framework is motivated by the observation that salient and
source-supported information often recurs across multiple decoding
trajectories, whereas unsupported or weakly grounded content is typically less
stable. Sentence-level selection can therefore combine complementary
information from different generations while filtering repetitive or
low-quality candidates. This design makes factuality, diversity, and budget
constraints explicit at selection time rather than leaving them implicit in
autoregressive decoding.

We instantiate the framework with a unified candidate-utility function and a
DPP-inspired log-determinant selector that promotes high-quality and diverse
subsets. The framework is modular with respect to both components: the
generator may be an encoder--decoder summarization model or an
instruction-tuned LLM, while the selector may be implemented using MMR,
integer programming, or the proposed determinant-based objective. Because the
similarity matrix is derived from heuristic redundancy scores, we use the
log-determinant criterion as a deterministic diversity-aware objective rather
than as a probabilistic DPP.

We evaluate the framework on the single-document CNN/DailyMail benchmark
\cite{nallapati-etal-2016-abstractive}, the multi-document Multi-News
benchmark \cite{fabbri-etal-2019-multi}, and the more recent FaithBench and
TofuEval factuality-oriented benchmarks
\cite{bao-etal-2025-faithbench,tang-etal-2024-tofueval}.
Across generators and datasets, global sentence selection consistently
improves automatic factuality and source-grounding metrics, with particularly
large gains in multi-document summarization. These improvements are
accompanied by lower ROUGE and BERTScore, revealing a clear trade-off between
source grounding and reference overlap. A blind human evaluation further
shows that selected summaries are preferred more frequently and receive higher
ratings for factual consistency, relevance, clarity, and conciseness, although
direct generation retains a small advantage in coherence.

Our contributions are threefold:
\begin{itemize}
    \item We propose a modular generation-and-selection framework for
    sentence-budget-constrained abstractive summarization that operates
    without retraining the underlying generator.

    \item We formulate sentence-level content planning as a combinatorial
    selection problem that combines relevance, factuality, and redundancy,
    and instantiate it with a DPP-inspired determinant-based
    quality--diversity objective.

    \item We provide extensive automatic, statistical, ablation, and human
    evaluation across single-document, multi-document, and
    factuality-oriented benchmarks, demonstrating improved source grounding
    together with an explicit reference-overlap trade-off.
\end{itemize}

\section{Related Work}

Our work connects three lines of research: neural abstractive summarization,
faithfulness-oriented generation, and combinatorial sentence selection.
Pretrained encoder--decoder models and instruction-tuned LLMs have substantially
improved summarization fluency, but remain sensitive to long contexts, input
ordering, and user-specified constraints such as length and topic
\cite{zhang-etal-2024-benchmarking,ravaut-etal-2024-context,
chhabra-etal-2024-revisiting,liu-etal-2024-benchmarking}. Recent modular
approaches decompose summarization into extraction, generation, selection, or
rewriting stages \cite{guan-padmakumar-2023-extract,
le-luu-2024-extractive}. In contrast, we treat the generator as a source of
multiple abstractive hypotheses and perform global sentence-level selection
without retraining or an additional rewriting stage.

Faithfulness has been addressed through factuality-aware decoding,
self-consistency, post-editing, refinement, and reranking
\cite{li-etal-2024-improving-faithfulness,chae-etal-2024-mitigating,
wan-etal-2025-mamm}. Related reranking methods such as SimCLS,
SummaReranker, and lightweight generation rerankers select one complete
generated hypothesis from a candidate set
\cite{liu-liu-2021-simcls,ravaut-etal-2022-summareranker,
jain-etal-2024-lightweight}. 
Recent factuality evaluators and benchmarks, including AlignScore,
MiniCheck, FaithBench, and FaithLens, provide complementary mechanisms for
assessing source grounding and identifying unsupported content
\cite{zha-etal-2023-alignscore,tang-etal-2024-minicheck,
bao-etal-2025-faithbench,si-etal-2026-faithlens}. Our method instead decomposes multiple generated
summaries into reusable sentence candidates and recombines complementary
content under explicit factuality, redundancy, and budget constraints. This
enables factuality to influence the final content plan directly rather than
serving only as a summary-level reranking or post-hoc correction signal.

Combinatorial optimization has long been used in extractive and
multi-document summarization through MMR, integer programming, submodular
maximization, and determinantal point processes
\cite{10.1145/290941.291025,gillick-favre-2009-scalable,
lin-bilmes-2011-class,10.1561/2200000044}. These methods typically select
sentences from the source document. Our framework transfers this global
selection perspective to language-model-generated abstractive candidates,
jointly incorporating source relevance, consistency, redundancy, and
an explicit sentence budget. A broader discussion of generation-based
summarization, hallucination reduction, and optimization-based selection is
provided in Appendix~\ref{sec:appendix_related_work}.

\section{Method}

\begin{figure*}[t]
\centering
\includegraphics[width=\textwidth]{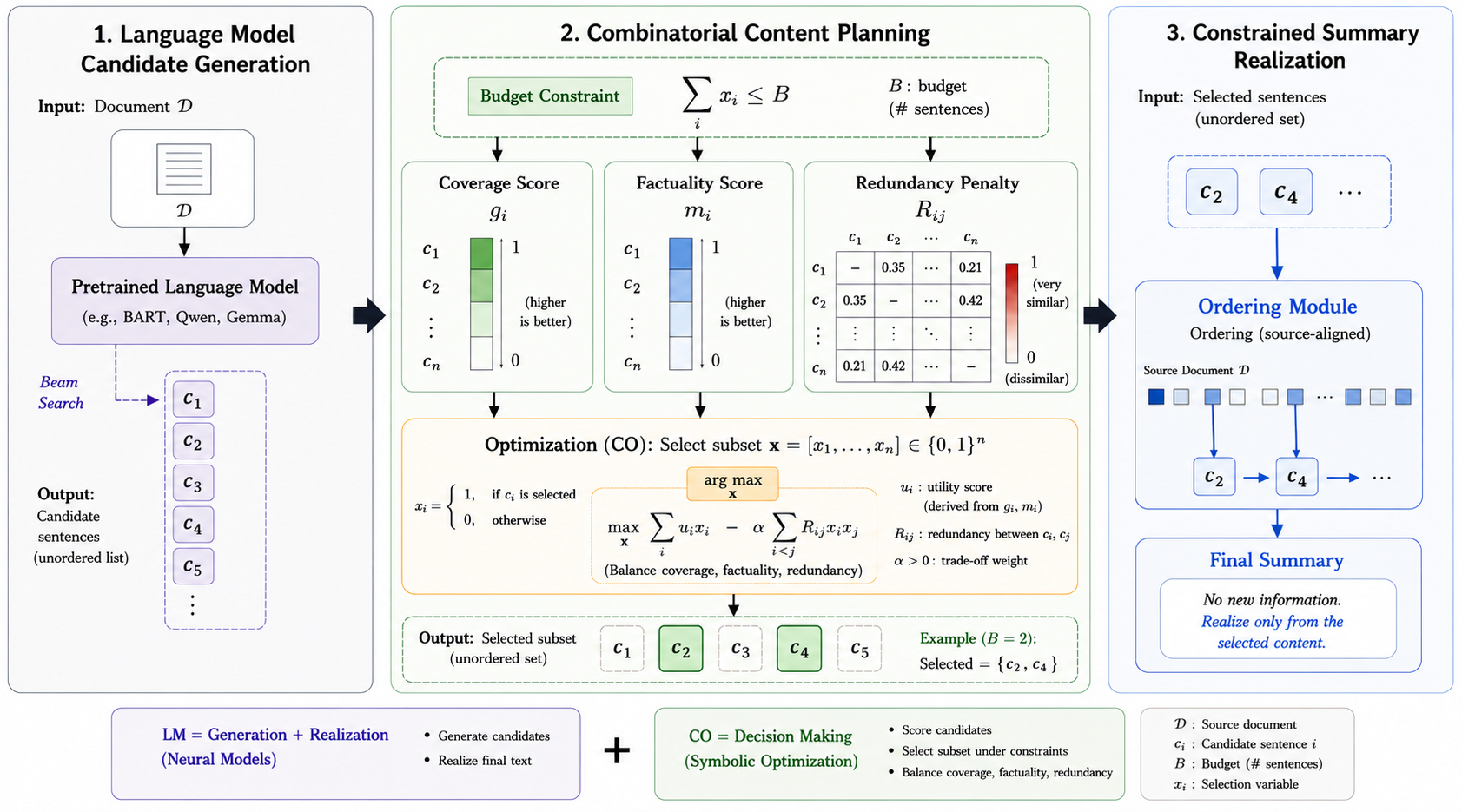}
\caption{
Overview of the proposed framework. A pretrained generator first produces
multiple candidate summaries, which are decomposed into sentence-level
candidates. A combinatorial selector then chooses a subset under a sentence
budget by balancing coverage, factuality, and redundancy. Finally, the selected
sentences are ordered according to the source document and concatenated to form
the final summary.
}
\label{fig:pipeline}
\end{figure*}

\subsection{Problem Definition and Framework Overview}

Given an input document or document set $\mathcal{D}$, the goal of abstractive
summarization is to produce a concise summary $\hat{y}$ that captures salient
information while remaining consistent with the source. Conventional systems
typically generate $\hat{y}$ through a single autoregressive decoding
trajectory. Although this formulation produces fluent text, it provides only
indirect control over which content is selected, how much information is
included, and whether different parts of the summary are redundant.

We instead formulate summarization as a generation-and-selection problem over a
finite candidate space. As illustrated in Figure~\ref{fig:pipeline}, the
framework consists of three stages. First, a pretrained generator $G$ produces
multiple candidate summaries for the same input. Second, the generated
summaries are decomposed into sentence-level candidates, which are scored for
coverage, factuality, and pairwise redundancy. Third, a combinatorial selector
chooses a subset of candidates under an explicit sentence budget, after which
the selected sentences are ordered and concatenated to form the final summary.

Formally, the generator produces
$
Y=\{y_1,y_2,\ldots,y_p\},
$
where each $y_j$ corresponds to a distinct decoding trajectory. The generated
summaries are segmented into sentences and aggregated into a candidate pool
$
\mathcal{C}=\{c_1,c_2,\ldots,c_n\}.
$
Exact duplicates are removed before optimization.

The final summary is constructed by selecting a subset
$A\subseteq\mathcal{C}$ under a sentence-level budget:
\begin{equation}
A^\ast
=
\arg\max_{A\subseteq\mathcal{C}}
F(A)
\quad
\text{s.t.}
\quad
|A|\leq B,
\label{eq:general_objective}
\end{equation}
where $B$ is the maximum number of selected sentences and $F(A)$ measures the
quality of a candidate subset. The objective favors subsets that cover salient
source information, contain source-supported claims, and avoid semantic
repetition.

This formulation separates fluent text generation from global content
selection. The generator explores multiple possible summaries, while the
selector compares candidate content across decoding trajectories and imposes
explicit constraints at the subset level. Because the selector operates over
generated sentences rather than source sentences, the framework retains the
abstractive capability of pretrained models while allowing global content
planning without retraining the generator.

\subsection{Candidate Generation}

The candidate-generation stage converts the open-ended output space of a
language model into a finite set of reusable sentence candidates. Rather than
committing to a single decoded summary, we generate multiple hypotheses using
beam search or stochastic decoding. Different trajectories may emphasize
different facts, paraphrases, or content combinations, thereby exposing a
broader set of possible summary units to the downstream selector.

This design is motivated by the observation that salient and source-supported
information often appears across multiple decoding trajectories, whereas
weakly supported or hallucinated content is less stable. The framework does not
assume that repeated content is necessarily correct; instead, it treats the
candidate pool as a search space whose elements are subsequently evaluated
against the source.

Each generated summary $y_j$ is segmented into sentences. The resulting
sentences are pooled and deduplicated to construct $\mathcal{C}$. We retain the
originating summary and within-summary position of each candidate for analysis
and realization, although subset selection is performed over the deduplicated
pool.

Unlike summary-level reranking, which selects one complete hypothesis from
$Y$, our framework treats generated sentences as reusable content units. This
allows the final summary to combine complementary information from multiple
candidate summaries under a shared budget. The generation stage is independent
of the selector and can therefore be used with encoder--decoder summarization
models or instruction-tuned LLMs without architectural modification.

\subsection{Candidate Scoring}

For each candidate sentence, we compute two sentence-level scores and one
pairwise score. The sentence-level scores measure coverage and factuality,
whereas the pairwise score estimates redundancy between candidates.

\paragraph{Coverage}

The coverage score measures how well a candidate captures salient information
from the source:
\begin{equation}
g_i=\phi(\mathcal{D},c_i),
\label{eq:coverage}
\end{equation}
where $\phi$ is a relevance function comparing candidate $c_i$ with
$\mathcal{D}$. Depending on the implementation, $\phi$ may use lexical overlap,
semantic similarity, contextual embeddings, or a learned relevance estimator.
Higher values indicate that the candidate is more representative of the source
content.

\paragraph{Factuality}

To encourage source grounding, we compute a sentence-level factuality score:
\begin{equation}
m_i
=
P_{\mathrm{fact}}
\bigl(\mathrm{consistent}\mid\mathcal{D},c_i\bigr),
\label{eq:factuality}
\end{equation}
where $P_{\mathrm{fact}}$ estimates whether the claim expressed by $c_i$ is
supported by the source. Candidates judged to be unsupported or weakly
grounded therefore receive lower utility during selection.

Coverage and factuality capture complementary properties. A sentence may be
well supported but peripheral to the main topic, or highly relevant but contain
an unsupported detail. We combine both signals into a unified sentence-level
utility:
\begin{equation}
u_i
=
\lambda_{\mathrm{cov}}g_i
+
\lambda_{\mathrm{fact}}m_i,
\label{eq:utility}
\end{equation}
where $\lambda_{\mathrm{cov}}$ and $\lambda_{\mathrm{fact}}$ control the
relative importance of relevance and source grounding.

\paragraph{Redundancy}

Sentence-level utility alone may favor several candidates expressing the same
fact. We therefore compute a pairwise redundancy matrix
$R\in\mathbb{R}^{n\times n}$:
\begin{equation}
R_{ij}=\psi(c_i,c_j),
\label{eq:redundancy}
\end{equation}
where $\psi$ measures lexical or semantic similarity between candidates
$c_i$ and $c_j$. Large values indicate that the two candidates convey
substantially overlapping content. Before optimization, the scoring components
are normalized within each candidate pool so that coverage, factuality, and
redundancy contribute on comparable scales.

\subsection{Budget-Constrained Selection}

Let
\[
\mathbf{z}=[z_1,\ldots,z_n]\in\{0,1\}^{n}
\]
denote the binary selection vector, where $z_i=1$ indicates that candidate
$c_i$ is selected. We first define a general quality--redundancy objective:
\begin{equation}
\begin{aligned}
\max_{\mathbf{z}\in\{0,1\}^{n}}
\quad &
\sum_{i=1}^{n} u_i z_i
-\alpha
\sum_{i<j} R_{ij} z_i z_j
\\
\text{s.t.}
\quad &
\sum_{i=1}^{n} z_i \leq B ,
\end{aligned}
\label{eq:selection_objective}
\end{equation}
where $\alpha\geq 0$ controls the strength of redundancy suppression. The
first term rewards individually informative and source-supported candidates,
whereas the second penalizes pairs that convey overlapping information.

Equation~\ref{eq:selection_objective} defines the general selection problem,
but the framework is not tied to a particular optimization algorithm. We
compare three selectors using identical candidate pools, utilities, pairwise
similarities, and sentence budgets. MMR performs greedy local selection by
balancing candidate utility against similarity to the current subset. The
integer-programming selector uses either hard pairwise redundancy constraints
or a soft linearized redundancy penalty, depending on the scoring
configuration. The DPP-inspired selector instead uses
a log-determinant criterion to evaluate candidate quality and subset diversity
jointly.

\paragraph{DPP-Inspired Selection}

The DPP-inspired selector uses the same candidate utilities and pairwise
redundancy scores as the other selectors, but replaces the additive pairwise
penalty in Eq.~\ref{eq:selection_objective} with a determinant-based subset
criterion. It is therefore an alternative instantiation of the general
selection framework rather than an exact reformulation of
Eq.~\ref{eq:selection_objective}.

For each candidate $c_i$, we derive a positive quality score from its
coverage--factuality utility:
\begin{equation}
q_i=\max(u_i,\epsilon_q),
\qquad
Q=\operatorname{diag}(q_1,\ldots,q_n),
\label{eq:dpp_quality}
\end{equation}
where $\epsilon_q$ prevents zero-quality entries. Let
$K\in\mathbb{R}^{n\times n}$ denote the similarity matrix obtained by scaling
the pairwise redundancy scores in Eq.~\ref{eq:redundancy} according to the
redundancy weight and clipping the resulting values to $[0,1]$. We construct
\begin{equation}
L=QKQ+\epsilon I,
\label{eq:dpp_kernel}
\end{equation}
where $\epsilon$ provides numerical regularization. The diagonal
quality terms favor candidates with high coverage--factuality utility, while
the log-determinant criterion is used to discourage subsets containing highly
similar candidates.

For a target sentence budget $B$, the selector seeks
\begin{equation}
\begin{aligned}
A^\ast
=
\arg\max_{A\subseteq\mathcal{C}}
\log\det(L_A)
\quad
\\
\text{s.t.}
|A|=\min(B,|\mathcal{C}|),
\end{aligned}
\label{eq:dpp_objective}
\end{equation}
where $L_A$ denotes the principal submatrix indexed by $A$. Unlike the general
constraint in Eq.~\ref{eq:selection_objective}, the implemented selector fills
the available sentence budget whenever the candidate pool contains at least
$B$ candidates.

Exact optimization is combinatorial, so we use greedy log-determinant
maximization. Starting from the empty set, the selector evaluates every
remaining candidate and adds the one producing the largest valid value of
$\log\det(L_{A\cup\{i\}})$. A candidate addition is valid only when the
corresponding determinant has positive sign. The process continues until the
target number of candidates has been selected.

If no remaining candidate yields a valid determinant, the selector adds the
candidate with the largest remaining diagonal entry of $L$. If the greedy
procedure encounters a numerical exception, it instead selects the candidates
with the highest quality scores. When $|\mathcal{C}|\leq B$, all candidates
are retained.

Because preprocessing does not guarantee that the heuristic similarity matrix
$K$ is positive semi-definite, the resulting $L$ need not define a valid DPP
kernel. We therefore use the log-determinant criterion as a deterministic
diversity heuristic and refer to the method as \emph{DPP-inspired}, rather than
as a probabilistic DPP.

\subsection{Summary Realization}

The selected subset $A^\ast$ is unordered, but a readable summary requires a
coherent sentence sequence. For each selected candidate $c_i$, we identify the
most semantically similar sentence in the source and assign the corresponding
source position $p(c_i)$. The selected candidates are then sorted by
$p(c_i)$ and concatenated to form the final summary $\hat{y}$.

This source-aligned ordering preserves the approximate information flow of the
input document and avoids introducing an additional generation step. We do not
apply neural rewriting after selection because rewriting could modify the
selected factual content or violate the sentence budget. The final output
therefore directly realizes the optimized subset, although combining sentences
from different decoding trajectories may occasionally introduce minor
discourse or stylistic inconsistencies.

\newcommand{\statval}[2]{%
  #1{\scriptsize\,${\pm}$#2}\textsuperscript{\(\ddag\)}%
}
\newcommand{\statns}[2]{%
  #1{\scriptsize\,${\pm}$#2}%
}
\newcommand{\best}[1]{%
  \textbf{#1}%
}
\newcommand{\beststat}[2]{%
  \textbf{#1{\scriptsize\,${\pm}$#2}}\textsuperscript{\(\ddag\)}%
}

\begin{table*}[t]
\centering
\small
\renewcommand{\arraystretch}{1.08}
\setlength{\tabcolsep}{4.0pt}

\resizebox{\linewidth}{!}{%
\begin{tabular}{@{}llcccccccccc@{}}
\toprule
\textbf{Dataset}
& \textbf{Method}
& \multicolumn{4}{c}{\textbf{ROUGE} ($\uparrow$)}
& \multicolumn{1}{c}{\textbf{BERTScore} ($\uparrow$)}
& \multicolumn{5}{c}{\textbf{Faithfulness} ($\uparrow$)}
\\
\cmidrule(lr){3-6}
\cmidrule(lr){7-7}
\cmidrule(l){8-12}
&
& \textbf{R-1}
& \textbf{R-2}
& \textbf{R-L}
& \textbf{R-Lsum}
& \textbf{F1}
& \textbf{FactCC}
& \textbf{MiniCheck}
& \textbf{AlignScore}
& \textbf{FactKB}
& \textbf{FaithLens}
\\
\midrule

\multirow{12}{*}{CNN/DM}
& BART
& 44.09 & 21.13 & 30.68 & 41.05
& 88.22 & 75.91 & 94.96 & 91.55 & 98.47 & 98.02
\\

& FactEdit
& 42.53 & 20.48 & 39.74 & 40.76
& 88.17 & 76.03 & 94.82 & 91.46 & 98.68 & 97.85
\\

& SimCLS
& 46.67 & 22.15 & 43.54 & 43.09
& 66.14 & 59.49 & 83.34 & 82.37 & 80.75 & 95.74
\\

& BRIO-Mul
& \best{47.78}
& \best{23.55}
& \best{44.57}
& \best{44.12}
& 87.84 & 54.06 & 84.67 & 74.48 & 85.20 & 94.23
\\

& BRIO-Ctr
& 47.28 & 22.93 & 44.15 & 44.10
& \best{89.11}
& 62.62 & 88.42 & 84.98 & 78.89 & 96.63
\\

& EFactSum
& 44.37 & 21.28 & 40.92 & 41.07
& 88.36 & 60.74 & 88.33 & 82.02 & 91.38 & 97.38
\\

& SummaReranker
& 46.62 & 22.39 & 43.59 & 43.80
& 88.47 & 68.17 & 93.81 & 88.27 & 98.10 & 97.68
\\

& Llama-3-8B
& 38.98 & 14.82 & 24.20 & 35.28
& 87.72 & 46.78 & 76.00 & 75.08 & 98.51 & 99.09
\\

& Qwen3.5-9B
& 34.88 & 10.91 & 20.33 & 30.77
& 86.96 & 40.16 & 56.82 & 65.17 & 98.12 & 98.69
\\

& Gemma-4-E4B-it
& 30.51 & 6.41 & 18.07 & 27.40
& 86.19 & 53.99 & 69.49 & 71.32 & 89.59 & 75.80
\\

\cmidrule(lr){2-12}

& BART+DPP
& \statval{39.18}{0.17}
& \statval{17.76}{0.16}
& \statval{26.23}{0.17}
& \statval{36.30}{0.17}
& \statval{87.23}{0.03}
& \beststat{79.93}{0.43}
& \beststat{97.73}{0.16}
& \beststat{94.63}{0.21}
& \beststat{99.46}{0.21}
& \beststat{99.63}{0.25}
\\

& Llama-3-8B+DPP
& \statval{26.40}{0.18}
& \statval{5.15}{0.14}
& \statval{16.39}{0.13}
& \statval{22.91}{0.16}
& \statval{85.06}{0.04}
& \statval{51.57}{0.64}
& \statval{87.56}{0.39}
& \statval{79.56}{0.40}
& \statns{98.55}{0.21}
& \statns{98.89}{0.25}
\\

\midrule

\multirow{9}{*}{Multi-News}
& PRIMERA
& 41.10 & 15.88 & 22.17 & 37.30
& 86.89 & 37.40 & 63.06 & 53.30 & 94.36 & 64.46
\\

& SimCLS
& 45.78
& \best{17.31}
& \best{23.14}
& 41.59
& 85.68 & 32.47 & 54.35 & 48.19 & 57.35 & 58.32
\\

& BRIO-Ctr
& \best{46.04}
& 16.90
& 22.73
& \best{41.78}
& 85.61 & 30.39 & 51.50 & 46.01 & 53.81 & 53.18
\\

& SummaReranker
& 44.56 & 16.72 & 22.61 & 40.45
& \best{87.14}
& 35.18 & 61.49 & 51.88 & 94.72 & 60.26
\\

& Llama-3-8B
& 38.84 & 8.68 & 18.16 & 34.81
& 85.42 & 38.35 & 60.46 & 59.04 & 90.99 & 68.52
\\

& Qwen3.5-9B
& 38.31 & 6.74 & 15.96 & 34.05
& 84.82 & 42.01 & 48.40 & 53.08 & 86.98 & 49.20
\\

& Gemma-4-E4B-it
& 35.44 & 6.53 & 16.27 & 31.86
& 84.89 & 48.49 & 66.79 & 66.35 & 87.68 & 57.93
\\

\cmidrule(lr){2-12}

& PRIMERA+DPP
& \statval{36.69}{0.24}
& \statval{11.86}{0.18}
& \statval{19.56}{0.17}
& \statval{32.79}{0.23}
& \statval{84.49}{0.05}
& \statval{45.68}{0.74}
& \beststat{85.38}{0.56}
& \statval{71.27}{0.55}
& \statval{97.54}{0.44}
& \statval{79.70}{1.45}
\\

& Llama-3-8B+DPP
& \statval{32.24}{0.20}
& \statval{5.42}{0.09}
& \statval{15.93}{0.09}
& \statval{28.18}{0.18}
& \statval{83.59}{0.04}
& \beststat{53.67}{0.80}
& \statval{83.15}{0.56}
& \beststat{72.56}{0.53}
& \beststat{98.29}{0.57}
& \beststat{95.14}{1.30}
\\

\bottomrule
\end{tabular}%
}

\caption{
Main results on CNN/DailyMail and Multi-News. Values following $\pm$ denote
half-widths of the 95\% paired-bootstrap confidence intervals for the
\texttt{+DPP}-minus-baseline differences. \ddag{} indicates
$p_{\mathrm{Holm}}<0.01$ under a two-sided paired sign-flip permutation test
with Holm correction over the ten reported metrics. Unmarked
\texttt{+DPP} results are not statistically significant after correction.
Bold values indicate the highest score for each metric within each dataset.
Complete paired differences, confidence intervals, and adjusted $p$ values
are reported in the appendix.
}
\label{tab:main_results}
\end{table*}

\section{Results}

Detailed experimental settings, evaluation protocols, complete statistical
results, and additional analyses are provided in
Appendix~\ref{sec:appendix_experimental}.

\subsection{Main Results}

Table~\ref{tab:main_results} compares the proposed DPP-based selection
framework with direct-generation baselines, prior summarization systems,
faithfulness-oriented methods, and instruction-tuned LLMs on CNN/DailyMail and
Multi-News.

Overall, DPP-based selection consistently improves factuality-oriented metrics
while reducing reference-overlap scores. On CNN/DailyMail,
\texttt{BART+DPP} achieves the strongest factuality performance across all five
reported metrics, obtaining FactCC $79.93$, MiniCheck $97.73$, AlignScore
$94.63$, FactKB $99.46$, and FaithLens $99.63$. Relative to direct
\texttt{BART}, the largest gains occur on FactCC and AlignScore, indicating that
sentence-level selection improves both entailment-based and semantic-alignment
measures of source grounding.

The gains are more pronounced on Multi-News. Compared with direct
\texttt{PRIMERA}, \texttt{PRIMERA+DPP} improves MiniCheck from $63.06$ to
$85.38$, AlignScore from $53.30$ to $71.27$, and FaithLens from $64.46$ to
$79.70$. Similarly, \texttt{Llama-3-8B+DPP} improves FactCC from $38.35$ to
$53.67$, MiniCheck from $60.46$ to $83.15$, AlignScore from $59.04$ to
$72.56$, FactKB from $90.99$ to $98.29$, and FaithLens from $68.52$ to
$95.14$. These larger improvements suggest that global sentence selection is
particularly effective when salient information and redundant content are
distributed across multiple source documents.

Nearly all paired differences are statistically significant after Holm
correction. The only unmarked comparisons are FactKB and FaithLens for
\texttt{Llama-3-8B+DPP} on CNN/DailyMail. Complete paired differences,
confidence intervals, and adjusted $p$ values are reported in the appendix.

The factuality gains are accompanied by lower ROUGE and BERTScore. This
trade-off reflects the selection objective, which prioritizes source-supported
and non-redundant content rather than lexical agreement with a single reference
summary. Reference overlap and factual grounding should therefore be
interpreted as complementary dimensions of summary quality rather than
interchangeable measures.

\subsection{Additional Faithfulness Benchmarks}

Table~\ref{tab:faithfulness_benchmarks} evaluates the framework on FaithBench
and TofuEval, which provide complementary tests of factual consistency beyond
the standard CNN/DailyMail and Multi-News settings.

On FaithBench, \texttt{BART+DPP} improves over direct \texttt{BART} on every
metric, increasing FactCC from $80.13$ to $89.25$, MiniCheck from $86.62$ to
$92.89$, AlignScore from $88.92$ to $93.94$, FactKB from $97.25$ to $99.90$,
and FaithLens from $86.67$ to $94.67$. The Llama-based selector also produces
substantial gains, particularly on MiniCheck, AlignScore, FactKB, and
FaithLens.

A similar pattern appears on TofuEval. \texttt{BART+DPP} achieves the strongest
overall factuality performance, including MiniCheck $96.47$ and AlignScore
$90.17$. \texttt{Llama-3-8B+DPP} likewise improves all five metrics over its
direct-generation baseline. FaithLens reaches $100.00$ for both DPP systems on
this benchmark, making the remaining factuality metrics important for
distinguishing their performance.

These results show that the source-grounding improvements extend beyond the
two standard news benchmarks and are observed across both task-specific
summarization models and instruction-tuned LLMs.

\begin{table}[t]
\centering
\small
\renewcommand{\arraystretch}{1.05}
\setlength{\tabcolsep}{3.5pt}
\resizebox{\columnwidth}{!}{%
\begin{tabular}{llccccc}
\toprule
\textbf{Dataset} & \textbf{Method}
& \textbf{FactCC}
& \textbf{MiniCheck}
& \textbf{AlignScore}
& \textbf{FactKB}
& \textbf{FaithLens}\\
\midrule

\multirow{10}{*}{FaithBench}
 & BART             & 80.13 & 86.62 & 88.92 & 97.25 & 86.67 \\
 & PRIMERA          & 34.11 & 31.22 & 33.44 & 81.85 & 28.00 \\
 & Llama-3-8B       & 37.25 & 47.54 & 55.76 & 60.98 & 66.67 \\
 & Qwen3.5-9B       & 38.11 & 29.81 & 41.48 & 66.64 & 40.00 \\
 & Gemma-4-E4B-it   & 49.13 & 54.02 & 59.74 & 60.71 & 53.33 \\
 & SimCLS           & 29.45 & 30.11 & 28.42 & 37.52 & 40.00 \\
 & BRIO-Ctr         & 30.23 & 28.57 & 27.94 & 36.40 & 44.00 \\
 & SummaReranker    & 29.27 & 34.06 & 34.66 & 74.76 & 45.33 \\
\cmidrule(l){2-7}
& BART+DPP      & \textbf{89.25} & \textbf{92.89} & \textbf{93.94} & \textbf{99.90} & \textbf{94.67}\\
& Llama-3-8B+DPP & 42.43 & 73.34 & 70.86 & 90.83 & 84.00 \\

\midrule

\multirow{10}{*}{TofuEval}
& BART             & 59.15 & 86.41 & 81.18 & 94.01 & 98.00 \\
& PRIMERA          & 35.82 & 44.11 & 41.15 & 92.52 & 33.00 \\
& Llama-3-8B       & 39.38 & 56.31 & 54.61 & 84.14 & 90.00 \\
& Qwen3.5-9B       & 34.09 & 37.65 & 44.42 & 78.82 & 79.00 \\
& Gemma-4-E4B-it   & 49.51 & 67.47 & 65.09 & 76.47 & 80.00 \\
& SimCLS           & 32.50 & 46.30 & 41.10 & 41.80 & 40.00 \\
& BRIO-Ctr         & 33.15 & 44.49 & 40.03 & 42.70 & 36.00 \\
& SummaReranker    & 37.93 & 51.73 & 45.20 & 94.12 & 44.00 \\
\cmidrule(l){2-7}
& BART+DPP     & \textbf{66.37} & \textbf{96.47} & \textbf{90.17} & \textbf{96.48} & \textbf{100.00} \\
& Llama-3-8B+DPP & 52.79 & 83.91 & 70.97 & 95.63 & 100.00 \\
\bottomrule
\end{tabular}%
}
\caption{
Results on FaithBench and TofuEval. DPP-based selection improves factuality
metrics relative to the corresponding direct-generation backbone. Best results
within each dataset are shown in bold.
}
\label{tab:faithfulness_benchmarks}
\end{table}

\begin{table*}[htbp]
\centering
\small
\renewcommand{\arraystretch}{1.1}
\setlength{\tabcolsep}{4.5pt}
\resizebox{\linewidth}{!}{%
\begin{tabular}{@{} l l cccc c ccccc @{}}
\toprule
\textbf{Dataset} & \textbf{Method}
 & \multicolumn{4}{c}{\textbf{ROUGE} ($\uparrow$)}
 & \textbf{BERTScore} ($\uparrow$)
 & \multicolumn{5}{c}{\textbf{Faithfulness} ($\uparrow$)} \\
\cmidrule(lr){3-6} \cmidrule(lr){7-7} \cmidrule(l){8-12}
 & 
 & R-1 & R-2 & R-L & R-Lsum
 & F1
 & FactCC & MiniCheck & AlignScore & FactKB & FaithLens \\
\midrule

\multirow{6}{*}{CNN/DM}
 & BART+MMR            & 41.47 & 18.86 & 27.16 & 38.43 & 87.58 & 78.86 & 96.20 & 93.00 & 98.95 & 98.22 \\
 & BART+ILP             & 41.03 & 18.86 & 27.12 & 38.11 & 87.55 & 80.18 & 97.62 & 94.46 & 99.31 & 99.55 \\
 & BART+DPP             & 39.18 & 17.76 & 26.23 & 36.30 & 87.23 & 79.93 & 97.73 & 94.63 & 99.46 & 99.63\\
 \cmidrule(lr){2-12}
 & Llama-3-8B+MMR  & 28.51 & 5.88 & 17.08 & 24.81 & 85.51 & 50.32 & 85.81 & 78.60 & 98.28  & 98.68\\
 & Llama-3-8B+ILP  & 27.54 & 5.41 & 16.71 & 23.95 & 85.30 & 52.04 & 87.89 & 79.64 & 98.18  & 98.96\\
 & Llama-3-8B+DPP  & 26.40 & 5.15 & 16.39 & 22.91 & 85.06 & 51.57 & 87.56 & 79.56 & 98.55  & 98.89\\

\midrule

\multirow{6}{*}{Multi-News}
 & PRIMERA+MMR        & 38.87 & 13.27 & 20.30 & 35.04 & 85.06 & 45.23 & 83.85 & 69.89 & 96.49 & 74.14 \\
 & PRIMERA+ILP        & 37.26 & 12.18 & 19.73 & 33.36 & 84.62 & 45.77 & 85.39 & 71.27 & 97.40 & 78.87\\
 & PRIMERA+DPP        & 36.69 & 11.86 & 19.56 & 32.79 & 84.49 & 45.68 & 85.38 & 71.27 & 97.54 & 79.70\\
  \cmidrule(lr){2-12}
 & Llama-3-8B+MMR   & 32.77 & 5.60 & 16.02 & 28.70 & 83.72 & 53.46 & 83.03 & 72.39 & 98.14 & 94.88 \\
 & Llama-3-8B+ILP   & 32.99 & 5.64 & 16.02 & 28.91 & 83.77 & 53.43 & 83.30 & 72.46 & 98.10 & 95.11 \\
 & Llama-3-8B+DPP   & 32.24 & 5.42 & 15.93 & 28.18 & 83.59 & 53.67 & 83.15 & 72.56 & 98.29 & 95.14\\

\bottomrule
\end{tabular}%
}
\caption{
Comparison of MMR, ILP, and DPP under identical candidate pools, scoring
functions, and sentence budgets. MMR generally preserves more reference
overlap, whereas ILP and DPP provide stronger factuality-oriented performance.
}
\label{tab:ablation}
\end{table*}

\subsection{Comparison of Selection Strategies}

Table~\ref{tab:ablation} compares MMR, ILP, and DPP using identical candidate
pools, scoring functions, and sentence budgets. This comparison isolates the
effect of the selector from differences in candidate generation.

MMR generally preserves higher ROUGE and BERTScore, whereas ILP and DPP tend
to obtain stronger factuality-oriented results. For BART, DPP achieves the
highest MiniCheck, AlignScore, and FactKB, while ILP obtains the highest
FactCC. For the Llama-based systems, ILP performs slightly better on several
CNN/DailyMail factuality metrics, whereas DPP obtains the strongest FactKB. On
Multi-News, ILP and DPP perform similarly for PRIMERA, while DPP achieves the
highest FactCC, AlignScore, and FactKB for Llama-3-8B.

No selector dominates every metric. Instead, MMR, ILP, and DPP provide
different operating points within the same generation-and-selection framework:
MMR retains more reference overlap, while ILP and DPP place greater emphasis on
global factuality and diversity. This result supports the modularity of the
framework and shows that the observed gains do not arise solely from generating
a larger candidate pool.

\subsection{Human Evaluation}

\begin{table}[t]
\centering
\small
\renewcommand{\arraystretch}{1.08}
\setlength{\tabcolsep}{3.8pt}

\begin{tabular*}{\columnwidth}{@{\extracolsep{\fill}}lrrrr@{}}
\toprule
\multicolumn{5}{@{}l}{\textbf{(a) Pairwise preference counts}} \\
\addlinespace[2pt]
\textbf{Backbone}
& \textbf{DPP}
& \textbf{Direct}
& \textbf{Same}
& \textbf{Unsure} \\
\midrule
BART       & \textbf{32} & 12 & 5  & 1 \\
Llama-3-8B & \textbf{29} & 14 & 5  & 2 \\
\midrule
\textbf{Overall} & \textbf{61} & 26 & 10 & 3 \\
\bottomrule
\end{tabular*}

\vspace{0.7em}

\begin{tabular*}{\columnwidth}{@{\extracolsep{\fill}}lrrr@{}}
\toprule
\multicolumn{4}{@{}l}{\textbf{(b) Mean quality ratings}} \\
\addlinespace[2pt]
\textbf{Dimension}
& \textbf{DPP}
& \textbf{Direct}
& \textbf{$\Delta$} \\
\midrule
Coherence
& 4.360
& \textbf{4.422}
& $-0.062$ \\
Consistency
& \textbf{4.466}
& 4.034
& $+0.432$ \\
Relevance
& \textbf{4.341}
& 3.708
& $+0.633$ \\
Clarity
& \textbf{4.379}
& 3.614
& $+0.765$ \\
Conciseness
& \textbf{4.218}
& 3.326
& $+0.892$ \\
\bottomrule
\end{tabular*}

\caption{
Human evaluation results. Panel~(a) reports pairwise preference counts over
100 comparisons. Panel~(b) reports mean ratings on a 1--5 scale, with
$\Delta$ computed as DPP-inspired selection minus direct generation. Bold
values indicate the preferred system or the higher mean rating.
}
\label{tab:human_eval_results}
\end{table}

To complement the automatic metrics, we conduct a blind pairwise evaluation on
100 CNN/DailyMail examples, including 50 comparisons for BART and 50 for
Llama-3-8B. Twenty annotators compare a direct-generation summary with its
DPP-selected counterpart. System identities and output order are hidden.
Annotators provide an overall preference and five-point ratings for
consistency, relevance, coherence, clarity, and conciseness. Full annotation
instructions are provided in Appendix~\ref{sec:human_protocol}.

Table~\ref{tab:human_eval_results} shows that DPP-selected summaries are
preferred in 61 comparisons, compared with 26 preferences for direct
generation; 10 pairs are judged equivalent and 3 are marked uncertain.
The preference pattern is consistent across both backbones, with DPP preferred
in 32 of 50 BART comparisons and 29 of 50 Llama-3-8B comparisons.

DPP-selected summaries also receive higher mean ratings for consistency,
relevance, clarity, and conciseness. Direct generation has a small numerical
advantage in coherence, suggesting that recombining sentences from multiple
candidate summaries may occasionally weaken discourse flow. Overall, the
human evaluation provides supporting evidence that the selection framework
improves perceived content quality and conciseness despite lower reference
overlap.

\subsection{Additional Analysis}

Appendix~\ref{sec:additional_results} reports sensitivity analyses for optimization weights, sentence
budget, and candidate-pool size. The sentence-budget analysis shows that the
DPP-inspired selector reliably satisfies the requested number of sentences
and retains its factuality advantage across budgets from two to five
sentences. Faithfulness-oriented weighting generally improves FactCC,
MiniCheck, and AlignScore, whereas diversity-oriented weighting preserves more
ROUGE-Lsum. Increasing the candidate pool further improves factuality,
particularly on Multi-News, but also increases generation and scoring cost and
strengthens the trade-off with reference overlap. A qualitative case study is provided in
Appendix~\ref{sec:appendix_case_study}.

\section{Conclusion}

We presented a modular generation-and-selection framework for
sentence-budget-constrained abstractive summarization. The framework generates
multiple candidate summaries and applies global sentence-level selection to
balance relevance, factuality, and redundancy without retraining the underlying
generator. Experiments across single-document, multi-document, and
factuality-oriented benchmarks show consistent improvements in automatic
source-grounding metrics, particularly for multi-document summarization, while
revealing a trade-off with reference-overlap scores. Human evaluation further
shows that selected summaries are preferred more frequently and receive higher
ratings for consistency, relevance, clarity, and conciseness. These
results demonstrate the potential of decoupling generation from global content
selection for controllable, factuality-oriented summarization.

\section{Limitations}

Despite achieving improvements in factual consistency and controllability, the proposed framework has several limitations.

First, the quality of the final summary remains fundamentally dependent on the quality of the generated candidate pool. Since the optimization stage only selects from generated candidates, factual information omitted by the underlying language model cannot be recovered during selection. Consequently, severe generation failures or systematic hallucinations in the candidate pool may limit downstream optimization effectiveness.

Second, the framework operates at the sentence level and does not explicitly
model discourse relations or cross-sentence dependencies. Although
source-aligned ordering preserves the approximate information flow of the
input, summaries assembled from different decoding trajectories may contain
stylistic inconsistencies or abrupt transitions. This limitation is also
reflected in the human evaluation, where direct generation obtains a small
numerical advantage in coherence. Moreover, because each comparison is
evaluated by a single annotator, inter-annotator agreement cannot be estimated.

Third, the framework uses automatic relevance and factuality scores during
selection. Improvements on related automatic evaluation metrics may therefore
partly reflect metric alignment and should not be interpreted as guarantees of
error-free factuality. The DPP-inspired selector also relies on greedy log-determinant maximization, which does not guarantee a globally optimal subset.
Finally, the current implementation enforces a sentence-count budget rather
than a finer-grained token- or word-level constraint.

Fourth, the framework introduces additional inference cost compared with direct generation baselines because multiple candidate summaries must be generated before optimization. The computational overhead increases with larger candidate pools and stronger factuality evaluation models.

Finally, although our evaluation spans news summarization and topic-focused dialogue summarization, it does not establish generalization to scientific, legal, clinical, or other long-form and highly abstractive domains. Evaluating the framework in these settings remains an important direction for future work.

\bibliography{custom}

\appendix

\section{Related Work}
\label{sec:appendix_related_work}

\subsection{Abstractive Summarization with Language Models}

Recent abstractive summarization research has increasingly shifted from task-specific encoder--decoder models to large language models (LLMs) and instruction-tuned generation systems \cite{wang2024surveying}. Prior work has benchmarked LLMs for news summarization, analyzed their zero-shot summarization behavior, and studied how they use long input contexts \cite{zhang-etal-2024-benchmarking,ravaut-etal-2024-context,chhabra-etal-2024-revisiting}. Other studies examine LLMs as teachers or references for training smaller summarization models, as well as their ability to follow user-specified constraints such as length, topic, or style \cite{liu-etal-2024-learning,liu-etal-2024-benchmarking,ryu-etal-2026-exploring}. Multi-document and long-document summarization remain especially challenging because models may fail to synthesize information across sources or may be sensitive to input ordering \cite{10.1162/tacl_a_00687,wang2024segmented,huang-etal-2024-embrace,kirstein-etal-2024-tell,sonowal-sadhu-2025-structure}. 

Our work differs from these generation-centered approaches by treating the language model as a candidate generator rather than the final decision maker. Instead of relying on a single decoded summary, we generate multiple candidate summaries, decompose them into sentence-level candidates, and apply a structured optimization layer to select a budget-constrained and faithful subset. This design is related to recent controllable and modular summarization studies, but our focus is on explicit post-generation content selection under global constraints \cite{jiang-dreyer-2024-ccsum,le-luu-2024-extractive,guan-padmakumar-2023-extract}.

\subsection{Faithfulness and Hallucination Reduction}

Faithfulness remains a central challenge in abstractive summarization. Earlier work showed that neural abstractive systems frequently generate unsupported content, motivating later studies on factual consistency evaluation and hallucination reduction \cite{maynez-etal-2020-faithfulness,wang-etal-2020-asking,fabbri-etal-2022-qafacteval}. Recent work has proposed stronger automatic evaluators and benchmarks, including alignment-based metrics, efficient grounded fact-checkers, and challenging hallucination benchmarks for summarization \cite{zha-etal-2023-alignscore,tang-etal-2024-minicheck,tam-etal-2023-evaluating,bao-etal-2025-faithbench}. Other studies analyze positional bias and context-use failures in long-form summarization, showing that LLMs may summarize the beginning and end of documents more faithfully than middle content \cite{ravaut-etal-2024-context,wan-etal-2025-positional}.

Beyond evaluation, recent methods attempt to reduce hallucination through decoding, self-consistency, active learning, refinement, or factuality-aware reranking \cite{li-etal-2024-improving-faithfulness,huangyw-etal-2025-dynamic,chae-etal-2024-mitigating,xia-etal-2024-hallucination,wan-etal-2025-mamm}. These methods typically modify generation, decoding, training, or post-hoc verification. Our method is complementary: we do not retrain the generator or ask the language model to rewrite the selected content. Instead, we use factuality as an explicit sentence-level scoring signal inside a budget-constrained selection problem, allowing hallucination reduction to be handled at the decision stage.

\subsection{Combinatorial Optimization for Summarization}

Combinatorial optimization has a long history in summarization, particularly for extractive and multi-document settings. Classical approaches formulate summarization as sentence selection under coverage, diversity, and budget constraints, using methods such as maximal marginal relevance, integer linear programming, submodular maximization, and determinantal point processes \cite{10.1145/290941.291025,gillick-favre-2009-scalable,lin-bilmes-2011-class,10.1561/2200000044}. These methods provide interpretable mechanisms for controlling redundancy and budget, but they are often applied to source sentences rather than language-model-generated abstractive candidates.

Recent work has renewed interest in modular generation, reranking, and candidate selection in the LLM era. Reranking methods select among generated candidates, while modular approaches decompose summarization into extraction, selection, and rewriting or refinement stages \cite{jain-etal-2024-lightweight,ri-etal-2025-reranking,guan-padmakumar-2023-extract,le-luu-2024-extractive}. Submodular and optimization-based selection has also appeared in related NLP settings such as in-context example selection, instruction tuning data mixture, and data-efficient learning \cite{ji-etal-2024-submodular,kumari-etal-2024-end,renduchintala-etal-2024-smart}. Our work builds on this line but differs by applying a diversity-aware combinatorial selection layer directly over LLM-generated sentence candidates, jointly incorporating coverage, factuality, redundancy, and an explicit budget constraint.

\section{Experimental Setup}
\label{sec:appendix_experimental}

\subsection{Datasets}

We evaluate the proposed framework on two standard news summarization datasets
and two factuality-oriented benchmarks. Together, they cover
single-document, multi-document, and topic-focused dialogue summarization, as
well as challenging hallucination settings involving modern LLMs.

\paragraph{CNN/DailyMail}

CNN/DailyMail \cite{nallapati-etal-2016-abstractive} is a widely used benchmark
for single-document news summarization. It contains news articles paired with
multi-sentence abstractive summaries written from article highlights. We use
the standard non-anonymized test split to evaluate summary quality and factual
consistency in a conventional single-document setting.

\paragraph{Multi-News}

Multi-News \cite{fabbri-etal-2019-multi} is a multi-document summarization
dataset containing clusters of related news articles paired with professionally
written summaries. Compared with CNN/DailyMail, it requires models to aggregate
information across multiple documents while controlling cross-document
redundancy and maintaining consistency over substantially longer inputs. We use
this dataset to evaluate whether global sentence selection is particularly
beneficial when relevant information is distributed across multiple sources.

\paragraph{FaithBench}

FaithBench \cite{bao-etal-2025-faithbench} is a hallucination benchmark designed
to evaluate summaries produced by modern LLMs. It contains 750
source--summary pairs constructed from 75 source passages and outputs from 10
LLMs spanning eight model families. The benchmark emphasizes challenging
examples on which widely used hallucination detectors disagree. Expert
annotators identify unsupported spans and categorize them as
\emph{benign}, \emph{questionable}, or \emph{unwanted}, in addition to fully
consistent cases. We use the source passages as summarization inputs and
evaluate whether the proposed selector improves source grounding on examples
associated with difficult and diverse hallucination patterns.

\paragraph{TofuEval}

TofuEval \cite{tang-etal-2024-tofueval} evaluates factual consistency in
topic-focused dialogue summarization. It is constructed from 100 long
dialogues, including 50 MediaSum interview transcripts and 50 MeetingBank city
council meetings. Three topics are identified for each dialogue, and five LLMs
generate a summary for each topic. After invalid outputs are removed, the
benchmark contains 1,479 summaries comprising 3,966 sentences. Professional
linguistic annotators provide sentence-level factuality labels and explanations
for unsupported statements, together with summary-level relevance and
completeness assessments. We use each dialogue--topic pair as a summarization
instance to examine whether the framework transfers beyond news articles to
long, multi-speaker, and conversational inputs.

CNN/DailyMail and Multi-News provide standard reference-based evaluation,
whereas FaithBench and TofuEval offer complementary tests of source grounding
under challenging LLM-generated hallucinations and domain transfer.

\subsection{Baselines}

We compare the proposed framework against representative pretrained summarization models, factuality-aware generation systems, reranking approaches, and recent instruction-tuned large language models.

\paragraph{Pretrained Summarization Models}

We include \textbf{BART} \cite{lewis-etal-2020-bart}, a widely adopted encoder--decoder summarization model, and \textbf{PRIMERA} \cite{xiao-etal-2022-primera}, a long-context summarization model designed for multi-document summarization.

\paragraph{Factuality-Aware Summarization Methods}

To evaluate faithfulness-oriented generation approaches, we compare against \textbf{FactEdit} \cite{balachandran-etal-2022-correcting}, which performs post-editing for factual correction, \textbf{BRIO-Mul} and \textbf{BRIO-Ctr} \cite{liu-etal-2022-brio}, which optimize generation quality through contrastive ranking objectives, and \textbf{EFactSum} \cite{dixit-etal-2023-improving}, which incorporates factuality-aware optimization into abstractive summarization.

\paragraph{Reranking and Candidate Selection Methods}

We additionally compare against \textbf{SimCLS} \cite{liu-liu-2021-simcls}, a contrastive learning-based reranking framework, and \textbf{SummaReranker} \cite{ravaut-etal-2022-summareranker}, which improves summarization quality through summary-level reranking.

\paragraph{Instruction-Tuned Large Language Models}

To evaluate the generality of the proposed framework across generation backbones, we further experiment with recent instruction-tuned large language models, including \textbf{Qwen3.5-9B} \cite{qwenteam2026qwen35omnitechnicalreport}, \textbf{Llama-3-8B} \cite{llama3modelcard}, and \textbf{Gemma-4-E4B-it} \cite{gemma2024, gemma4_e4b_it}.

\subsection{Implementation Details}

All experiments are implemented using \texttt{PyTorch} and the
\texttt{Hugging Face Transformers} library. For each input, the generation
model first produces multiple candidate summaries, which are segmented into
sentence-level candidates to construct the optimization pool
$\mathcal{C}$. Unless otherwise specified, all optimization-based methods
operate on the same cached candidate pools and sentence-level scores, ensuring
that comparisons among MMR, ILP, and DPP-inspired selection isolate the effect
of the selector.

\paragraph{Default Experimental Configuration}

The main optimization-based experiments use a candidate-generation width of
$12$ and a target sentence budget of $B=3$. All selector comparisons in
Table~\ref{tab:ablation} use identical candidate pools, scoring
functions, and sentence budgets. The sentence-budget analysis varies
$B\in\{2,3,4,5\}$ while holding the beam-12 candidate pools and candidate
scores fixed in Table~\ref{tab:budget_sweep_fully_faithful}. The candidate-pool analysis varies the generation width over
$\{4,8,12\}$, while the weight-sensitivity analysis uses beam size~$8$ and
the configurations reported in Table~\ref{tab:pool_size}.

Unless otherwise specified, the main experiments use the balanced weighting
configuration
\[
(w_{\mathrm{cov}},w_{\mathrm{fact}},w_{\mathrm{red}})
=
(0.33,0.33,0.34),
\]
where the three components control ROUGE-based coverage, MiniCheck-based
factuality, and pairwise redundancy, respectively.

\paragraph{Generation Models}

For CNN/DailyMail, we use
\texttt{facebook/bart-large-cnn}
\cite{lewis-etal-2020-bart}. The direct-generation baseline follows the
checkpoint's standard summarization configuration. The main
optimization-based experiments retain all generated
hypotheses for candidate-pool construction. Other decoding settings follow
the checkpoint configuration, including length normalization, no-repeat
trigram blocking, and early stopping.

For Multi-News, we use
\texttt{allenai/PRIMERA-multinews}
\cite{xiao-etal-2022-primera}. The direct-generation baseline follows the default decoding configuration provided by the checkpoint, whereas optimization runs generate a larger number of candidate summaries to improve candidate diversity.

For instruction-tuned language models, we use concise, dataset-specific
summarization prompts. The evaluated checkpoints are
\texttt{Qwen/Qwen3.5-9B}
\cite{qwenteam2026qwen35omnitechnicalreport},
\texttt{meta-llama/Llama-3.1-8B-Instruct}
\cite{llama3modelcard}, and
\texttt{google/gemma-4-E4B-it}
\cite{gemma2024,gemma4_e4b_it}. For compactness, we refer to
\texttt{meta-llama/Llama-3.1-8B-Instruct} as Llama-3-8B in the tables and
discussion. Direct-generation baselines use one returned summary, whereas the
main optimization-based runs retain $12$ generated hypotheses per input
before sentence-level selection.

For Multi-News inputs, source documents are concatenated in their original
dataset order using explicit document delimiters. Inputs are truncated only
when necessary to satisfy the maximum context length of the corresponding
generation model.

\paragraph{Candidate Pool Construction}

Generated summaries are segmented into sentence-level candidates using
standard sentence tokenization. Sentences from all retained hypotheses are
combined into a single pool, and exact duplicates are removed before scoring.
The resulting deduplicated pool is shared by all downstream selectors. We
retain the originating hypothesis and within-summary position of each
candidate for analysis and final summary realization.

\paragraph{Scoring Configuration}

Candidate coverage is computed using a ROUGE-based relevance score, while
sentence-level factuality is estimated using MiniCheck
\cite{tang-etal-2024-minicheck}. Pairwise redundancy is computed using
semantic similarity between candidate sentences. For each selected candidate,
source-aligned realization identifies its most semantically similar source
sentence and uses the corresponding source position for ordering.

Coverage, factuality, and redundancy scores are normalized independently
within each candidate pool. Sentence-level coverage and factuality scores are
normalized using min--max scaling. For redundancy, min--max normalization is
applied to the off-diagonal entries of the pairwise matrix, while diagonal
entries are retained as self-similarity values. This places the three scoring
components on comparable scales while preserving their relative ranking
within each candidate pool.

\paragraph{Optimization}

MMR greedily selects candidates by balancing normalized utility against
similarity to the current subset until the sentence budget is filled. The
integer-programming variants are implemented in PuLP and solved using the CBC
solver. The hard formulation selects exactly $B$ candidates and introduces a
mutual-exclusion constraint for each candidate pair whose redundancy exceeds
the configured threshold. The tri-metric formulation uses binary auxiliary
variables to linearize the pairwise redundancy penalty and permits
$1\leq\sum_i z_i\leq B$. In the reported tri-metric experiments, the
redundancy coefficient uses per-edge scaling:
\[
\alpha
=
\frac{w_{\mathrm{red}}}{\max(1,B-1)}.
\]

For the DPP-inspired selector, candidate utilities are clipped below at
$\epsilon_q=0.01$ to construct the diagonal quality matrix $Q$. Pairwise
redundancy scores are scaled according to $w_{\mathrm{red}}$ and clipped to
$[0,1]$ to form $K$. We then construct
\[
L=QKQ+10^{-6}I
\]
and apply greedy log-determinant maximization until the target budget is
filled. Candidate additions with non-positive determinant sign are treated as
invalid. If no valid candidate remains, the selector adds the candidate with
the largest remaining diagonal entry of $L$. A numerical exception triggers
fallback selection by descending candidate quality. When
$|\mathcal{C}|\leq B$, all candidates are retained.

\subsection{Evaluation Metrics}

We evaluate summarization quality using both generation-quality metrics and factual consistency metrics.

\paragraph{Generation Quality Metrics}

We report \textbf{ROUGE} \cite{lin-2004-rouge}, including ROUGE-1, ROUGE-2, ROUGE-L, and ROUGE-Lsum, to evaluate lexical overlap between generated summaries and reference summaries. We additionally report \textbf{BERTScore} \cite{zhang2019bertscore}, which measures semantic similarity using contextualized embeddings.

\paragraph{Faithfulness Metrics}

To evaluate factual consistency, we report several complementary faithfulness metrics. \textbf{FactCC} \cite{kryscinski-etal-2020-evaluating} evaluates whether generated summaries are entailed by the source document. \textbf{MiniCheck} \cite{tang-etal-2024-minicheck} provides efficient sentence-level factuality estimation for grounded generation. \textbf{AlignScore} \cite{zha-etal-2023-alignscore} measures semantic alignment between generated summaries and source documents using alignment-based scoring. We additionally report \textbf{FactKB} \cite{feng-etal-2023-factkb}, a knowledge-grounded factual consistency metric for abstractive generation. Finally, \textbf{FaithLens} \cite{si-etal-2026-faithlens} is an 8B-parameter explainable faithfulness evaluator that determines whether a generated claim is supported by the source document and produces a corresponding rationale for its prediction.

Together, these metrics provide a comprehensive evaluation of summary informativeness, semantic quality, and factual faithfulness.

\section{Case Study}
\label{sec:appendix_case_study}

\begin{table*}[t]
\centering
\footnotesize
\renewcommand{\arraystretch}{1.15}
\setlength{\tabcolsep}{5pt}

\begin{tabularx}{\textwidth}{>{\raggedright\arraybackslash\bfseries}p{0.16\textwidth} X}
\toprule
\rowcolor{gray!15}
\textbf{Component} & \textbf{Content} \\
\midrule

\rowcolor{gray!6}
Source Article
&
British intelligence agency GCHQ was ordered to destroy documents obtained from the unlawful interception of communications between Libyan dissident Sami al-Saadi and his lawyers. The Investigatory Powers Tribunal found that the intercepted material involved \textbf{legally privileged communications} and ordered the agency to destroy copies of the eavesdropped material. Al-Saadi claimed that the British government was involved in his \textbf{kidnapping} and return to Libya, where he said he was \textbf{tortured} by the Gaddafi regime. \\
\midrule

\rowcolor{gray!6}
Reference Summary
&
Sami al-Saadi's communications with his lawyer were \textbf{illegally intercepted}. The Investigatory Powers Tribunal ordered intelligence agencies to \textbf{destroy copies} of the intercepted communications. Mr.~al-Saadi claimed that the British government was complicit in \textbf{kidnapping} and returning him to Libya, where he was allegedly \textbf{tortured}. \\
\midrule

\rowcolor{gray!6}
Direct Baseline
&
A British watchdog ordered intelligence agencies to destroy illegally obtained information from communications between a Libyan dissident and his lawyers. GCHQ admitted intercepting confidential calls and emails involving Sami al-Saadi, violating legal professional privilege. The tribunal ruled that the dissident should not receive \textbf{compensation} for the violation of his privacy rights. GCHQ was instructed to destroy the intercepted documents to protect the dissident's privacy. \\
\midrule

\rowcolor{gray!6}
DPP Summary
&
A British watchdog ordered intelligence agencies to destroy illegally obtained information from a dissident's confidential communications with his lawyers. The intercepted material violated rules of \textbf{legal privilege}. The tribunal ordered the \textbf{destruction of the intercepted communications}. Sami al-Saadi also claimed that the British government was involved in his \textbf{kidnapping and torture} by the Libyan regime. \\
\midrule
\end{tabularx}

\caption{
Case study from CNN/DailyMail using \texttt{Llama-3.1-8B-Instruct}. Compared with direct generation, the proposed DPP-based selection framework better preserves reference-aligned factual structure, including the legal-privilege violation, destruction order, and kidnapping-and-torture allegation. The selected summary also achieves substantially higher factual consistency scores while improving reference alignment.
}
\label{tab:case_study_gchq}

\end{table*}

To better understand how combinatorial content planning affects the final summary, we present a representative example from the CNN/DailyMail test set using \texttt{meta-llama/Llama-3.1-8B-Instruct}. We compare the direct generation baseline with the proposed DPP-based sentence selection framework. Table~\ref{tab:case_study_gchq} presents the source article excerpt, the reference summary, both system outputs, and the corresponding quality and factuality metrics.

The example illustrates how post-generation sentence selection can recover source-supported information that is underemphasized or omitted in a single decoding trajectory, while simultaneously improving factual grounding. The direct generation baseline produces a fluent summary and captures the overall legal dispute, but allocates substantial space to compensation details and privacy-rights paraphrases that are not emphasized in the reference summary. In contrast, the DPP-selected summary more closely follows the reference structure by preserving several key factual elements, including the illegal interception, the legal-privilege violation, the destruction order, and the kidnapping-and-torture allegation.

This improvement is reflected in both overlap-based and factuality-oriented evaluation metrics. Compared with the direct generation baseline, the DPP-selected summary improves ROUGE-Lsum from $25.68$ to $41.86$, indicating substantially stronger alignment with the reference summary. More importantly, factual consistency metrics also improve consistently: AlignScore increases from $77.23$ to $86.39$, FactKB improves from $89.51$ to $99.32$, and MiniCheck increases from $95.22$ to $95.75$.

The example highlights the central intuition behind the proposed framework. Rather than relying on a single autoregressive generation trajectory, the combinatorial selector aggregates complementary factual units across multiple generated candidates under an explicit budget constraint. As a result, the final summary achieves stronger factual grounding and improved content coverage while avoiding redundant or weakly supported information.

\section{Additional Results}
\label{sec:additional_results}

This appendix provides sensitivity analyses
that complement the main findings.

\begin{table*}[t]
\centering
\small
\renewcommand{\arraystretch}{1.1}
\setlength{\tabcolsep}{4pt}

\begin{tabular}{llccccccc}
\toprule
\textbf{Dataset} & \textbf{Model}
& \textbf{Weight}
& \textbf{R-Lsum}
& \textbf{FactCC}
& \textbf{MiniCheck}
& \textbf{AlignScore}
& \textbf{FactKB}
& \textbf{FaithLens}\\
\midrule

\multirow{8}{*}{CNN/DM}
& \multirow{4}{*}{BART+DPP}
& Bal.   & 23.43 & 74.33 & 97.51 & 92.09 & \textbf{99.59} & \textbf{98.60} \\
&
& Cov.   & 21.01 & 73.25 & 96.96 & 91.45 & 99.46 & 98.40 \\
&
& Faith. & 23.12 & \textbf{74.67} & \textbf{97.63} & \textbf{92.60} & 99.50 & 98.40 \\
&
& Div.   & \textbf{26.24} & 73.37 & 97.35 & 91.99 & 99.35 & 98.40 \\
\cmidrule(lr){2-9}
& \multirow{4}{*}{Llama+DPP}
& Bal.   & 19.54 & 48.81 & 86.23 & 81.26 & 99.43 & 98.00 \\
&
& Cov.   & 19.00 & 46.57 & 80.98 & 78.96 & \textbf{99.52} & 98.00 \\
&
& Faith. & 19.02 & \textbf{50.01} & \textbf{87.81} & \textbf{82.00} & 99.43 & \textbf{98.60} \\
&
& Div.   & \textbf{20.48} & 48.42 & 85.55 & 80.87 & 99.22 & 98.00 \\

\midrule

\multirow{8}{*}{Multi-News}
& \multirow{4}{*}{PRIMERA+DPP}
& Bal.   & 33.43 & 42.31 & 79.37 & 65.59 & 97.80 & 78.40 \\
&
& Cov.   & 33.64 & 39.32 & 75.71 & 62.58 & \textbf{98.36} & \textbf{78.80} \\
&
& Faith. & 32.83 & \textbf{43.75} & \textbf{80.43} & \textbf{66.41} & 97.48 & 77.00 \\
&
& Div.   & \textbf{34.64} & 42.06 & 79.08 & 65.28 & 97.71 & 76.60 \\
\cmidrule(lr){2-9}
& \multirow{4}{*}{Llama+DPP}
& Bal.   & 30.41 & 52.60 & 78.22 & 68.94 & 97.98 & 93.20 \\
&
& Cov.   & 30.85 & 50.61 & 73.79 & 66.93 & 97.80 & 92.60 \\
&
& Faith. & 29.59 & \textbf{53.63} & \textbf{79.25} & \textbf{69.80} & \textbf{98.24} & \textbf{94.00} \\
&
& Div.   & \textbf{31.26} & 52.23 & 77.16 & 68.61 & 97.51 & 92.00\\

\bottomrule
\end{tabular}

\caption{
Trade-offs between coverage, faithfulness, and diversity under different optimization weight profiles.
All results use DPP selection with beam size $8$.
Weight abbreviations denote:
Bal. = $(0.33,0.33,0.34)$,
Cov. = $(0.60,0.20,0.20)$,
Faith. = $(0.20,0.60,0.20)$, and
Div. = $(0.20,0.20,0.60)$,
where the tuple corresponds to $(w_R,w_M,w_D)$ for ROUGE-based coverage, MiniCheck-based faithfulness, and redundancy/diversity control.
Best values within each model block are shown in bold.
}
\label{tab:tradeoff}
\end{table*}

\subsection{Sensitivity to Optimization Weights}

Table~\ref{tab:tradeoff} analyzes the effect of varying the relative weights
assigned to coverage, factuality, and diversity while keeping candidate
generation and DPP selection fixed. We compare balanced, coverage-heavy,
faithfulness-heavy, and diversity-heavy configurations.

Faithfulness-heavy weighting generally produces the strongest factuality
results. On CNN/DailyMail, the faithfulness-heavy setting improves
\texttt{Llama+DPP} from $48.81$ to $50.01$ on FactCC, from $86.23$ to
$87.81$ on MiniCheck, and from $81.26$ to $82.00$ on AlignScore relative to
balanced weighting. For \texttt{BART+DPP}, the same configuration obtains the
highest FactCC, MiniCheck, and AlignScore.

A similar pattern appears on Multi-News. For \texttt{PRIMERA+DPP},
faithfulness-heavy weighting improves FactCC from $42.31$ to $43.75$,
MiniCheck from $79.37$ to $80.43$, and AlignScore from $65.59$ to $66.41$.
For \texttt{Llama+DPP}, it improves FactCC from $52.60$ to $53.63$,
MiniCheck from $78.22$ to $79.25$, and AlignScore from $68.94$ to $69.80$.

Diversity-heavy weighting often yields the highest ROUGE-Lsum, whereas
faithfulness-heavy weighting generally achieves the strongest factuality
scores. This result further demonstrates that the framework supports explicit
control over the trade-off between reference coverage, source grounding, and
redundancy reduction.

\subsection{Effect of Sentence Budget and Output Length}
\label{sec:budget_analysis}

To assess budget controllability and determine whether the observed factuality
gains depend on a particular output length, we vary the target sentence budget
over $B\in\{2,3,4,5\}$. For each setting, we report average output length in
words, exact-budget adherence, reference-overlap scores, and
factuality-oriented metrics.

\begin{table*}[t]
\centering
\small
\renewcommand{\arraystretch}{1.08}
\setlength{\tabcolsep}{3.8pt}

\resizebox{\linewidth}{!}{%
\begin{tabular}{@{}llcccccccccc@{}}
\toprule
&
&
&
\multicolumn{2}{c}{\textbf{Output Control}}
&
\multicolumn{2}{c}{\textbf{Reference Overlap}}
&
\multicolumn{5}{c}{\textbf{Factuality}}
\\
\cmidrule(lr){4-5}
\cmidrule(lr){6-7}
\cmidrule(lr){8-12}

\textbf{Dataset}
& $\mathbf{B}$
& \textbf{Method}
& \textbf{Avg. words}
& \textbf{Hit (\%) $\uparrow$}
& \textbf{R-Lsum $\uparrow$}
& \textbf{BERTScore $\uparrow$}
& \textbf{FactCC $\uparrow$}
& \textbf{MiniCheck $\uparrow$}
& \textbf{AlignScore $\uparrow$}
& \textbf{FactKB $\uparrow$}
& \textbf{FaithLensFF $\uparrow$}
\\
\midrule

\multirow{8}{*}{\shortstack{CNN/DM\\(BART)}}
& \multirow{2}{*}{2}
& Direct
& 36.08 & 99.76
& \textbf{37.19} & \textbf{88.14}
& 74.76 & 95.24 & 91.85 & 98.51 & 99.46
\\
&
& +DPP
& 44.57 & \textbf{100.00}
& 32.65 & 86.84
& \textbf{80.94} & \textbf{98.01} & \textbf{94.99}
& \textbf{98.77} & \textbf{99.87}
\\
\cmidrule(lr){2-12}

& \multirow{2}{*}{3}
& Direct
& 53.44 & 94.26
& \textbf{40.35} & \textbf{88.22}
& 76.13 & 95.12 & 91.69 & 98.57 & 99.23
\\
&
& +DPP
& 62.78 & \textbf{100.00}
& 36.13 & 87.18
& \textbf{80.53} & \textbf{97.90} & \textbf{94.84}
& \textbf{99.18} & \textbf{99.74}
\\
\cmidrule(lr){2-12}

& \multirow{2}{*}{4}
& Direct
& 62.39 & 54.67
& \textbf{41.01} & \textbf{88.23}
& 76.08 & 95.02 & 91.62 & 98.48 & 98.51
\\
&
& +DPP
& 80.00 & \textbf{100.00}
& 36.66 & 87.23
& \textbf{79.93} & \textbf{97.73} & \textbf{94.63}
& \textbf{99.46} & \textbf{99.63}
\\
\cmidrule(lr){2-12}

& \multirow{2}{*}{5}
& Direct
& 64.48 & 13.73
& \textbf{41.04} & \textbf{88.22}
& 75.94 & 94.96 & 91.56 & 98.47 & 98.08
\\
&
& +DPP
& 96.85 & \textbf{100.00}
& 35.35 & 87.06
& \textbf{79.48} & \textbf{97.59} & \textbf{94.43}
& \textbf{99.44} & \textbf{99.50}
\\

\midrule

\multirow{8}{*}{\shortstack{Multi-News\\(PRIMERA)}}
& \multirow{2}{*}{2}
& Direct
& 52.64 & \textbf{100.00}
& \textbf{24.24} & \textbf{86.55}
& 23.81 & 52.46 & 41.97 & \textbf{91.75} & 81.00
\\
&
& +DPP
& 51.04 & \textbf{100.00}
& 19.85 & 84.62
& \textbf{47.48} & \textbf{92.86} & \textbf{78.76}
& 91.08 & \textbf{96.03}
\\
\cmidrule(lr){2-12}

& \multirow{2}{*}{3}
& Direct
& 74.92 & 99.36
& \textbf{30.38} & \textbf{86.78}
& 30.25 & 57.58 & 47.71 & 92.93 & 75.65
\\
&
& +DPP
& 72.00 & \textbf{100.00}
& 24.75 & 84.81
& \textbf{48.08} & \textbf{92.28} & \textbf{77.70}
& \textbf{93.27} & \textbf{93.72}
\\
\cmidrule(lr){2-12}

& \multirow{2}{*}{4}
& Direct
& 93.21 & 92.99
& \textbf{34.41} & \textbf{86.89}
& 33.83 & 60.63 & 51.07 & 93.66 & 71.06
\\
&
& +DPP
& 92.67 & \textbf{100.00}
& 28.13 & 84.84
& \textbf{47.96} & \textbf{91.32} & \textbf{76.56}
& \textbf{94.60} & \textbf{91.20}
\\
\cmidrule(lr){2-12}

& \multirow{2}{*}{5}
& Direct
& 103.47 & 66.36
& \textbf{36.42} & \textbf{86.91}
& 36.04 & 62.35 & 52.68 & 94.19 & 67.38
\\
&
& +DPP
& 113.80 & \textbf{100.00}
& 30.42 & 84.77
& \textbf{47.56} & \textbf{90.18} & \textbf{75.36}
& \textbf{95.60} & \textbf{88.28}
\\

\bottomrule
\end{tabular}%
}

\caption{
Sentence-budget and output-length analysis for direct generation and
DPP-inspired selection. Avg.\ words reports realized output length, and Hit
denotes the percentage of outputs containing exactly $B$ sentences.
FaithLensFF denotes the percentage of summaries for which every sentence is
classified as source-supported by FaithLens. Comparisons are matched by target
sentence budget rather than word count. Bold values indicate the better result
within each dataset, budget, and metric pair.}
\label{tab:budget_sweep_fully_faithful}
\end{table*}

Table~\ref{tab:budget_sweep_fully_faithful} shows that the DPP-inspired
selector satisfies the requested sentence budget for every evaluated example,
whereas direct generation becomes increasingly unreliable as $B$ grows. The
exact-budget hit rate of direct BART decreases from $99.76\%$ at $B=2$ to
$13.73\%$ at $B=5$, while direct PRIMERA decreases from $100.00\%$ to
$66.36\%$. In contrast, the DPP-inspired selector maintains a $100\%$ hit rate
across all budgets and both datasets.

Across budget settings, DPP-inspired selection consistently improves most
factuality-oriented metrics, including FactCC, MiniCheck, AlignScore, and the
percentage of fully faithful summaries, although these gains are accompanied
by lower ROUGE-Lsum and BERTScore. The source-grounding improvements therefore
persist across multiple sentence budgets rather than arising from a single
length configuration.

Because the constraint is defined over sentences rather than words, summaries
with the same target budget may still differ substantially in realized word
count. Notably, the DPP-inspired outputs are often longer than the corresponding
direct-generation outputs at the same value of $B$, particularly on
CNN/DailyMail. Their factuality improvements therefore cannot be attributed
simply to producing shorter summaries. Reporting both exact-budget adherence
and average word count provides a more complete view of controllability under
a sentence-level budget.

\subsection{Effect of Candidate Pool Size}

\begin{table*}[t]
\centering
\small
\renewcommand{\arraystretch}{1.1}
\setlength{\tabcolsep}{5pt}

\begin{tabular}{llccccccc}
\toprule
\textbf{Dataset} & \textbf{Method} & \textbf{Beam}
& \textbf{R-Lsum}
& \textbf{FactCC}
& \textbf{MiniCheck}
& \textbf{AlignScore}
& \textbf{FactKB}
& \textbf{FaithLens}
\\
\midrule

\multirow{6}{*}{CNN/DM}
& \multirow{3}{*}{BART}
& 4  &\textbf{41.05} & 75.91 & 94.96 & 91.55 & 98.47 & 98.02 \\
&
& 8  & 36.68 & 79.21 & 97.60 & 94.28 & 99.23 & 99.48 \\
&
& 12 & 36.30 & \textbf{79.93} & \textbf{97.73} & \textbf{94.63} & \textbf{99.46} & \textbf{99.63} \\
\cmidrule(lr){2-9}
& \multirow{3}{*}{Llama-3-8B}
& 4  & \textbf{35.28} & 46.78 & 76.00 & 75.08 & \textbf{98.51} & \textbf{99.09} \\
&
& 8  & 23.65 & 51.05 & 85.60 & 78.48 & 98.35 & 98.73 \\
&
& 12 & 22.91 & \textbf{51.57} & \textbf{87.56} & \textbf{79.56} & 98.50 & 98.89 \\
\midrule

\multirow{6}{*}{Multi-News}
& \multirow{3}{*}{PRIMERA}
& 4  & \textbf{37.30} & 37.37 & 62.26 & 52.24
& 94.31 & 64.46 \\
&
& 8  & 33.74 & 43.35 & 80.51 & 67.39 & \textbf{97.68} & 75.49 \\
&
& 12 & 32.79 & \textbf{45.68} & \textbf{85.38} & \textbf{71.27} & 97.54 & \textbf{79.70} \\
\cmidrule(lr){2-9}
& \multirow{3}{*}{Llama-3-8B}
& 4  & \textbf{34.81} & 38.35 & 60.46 & 59.04 & 90.99 & 68.52 \\
&
& 8  & 29.43 & 52.68 & 79.40 & 70.23 & 98.14 & 93.92 \\
&
& 12 & 28.18 & \textbf{53.67} & \textbf{83.15} & \textbf{72.56} & \textbf{98.29} & \textbf{95.14} \\
\bottomrule
\end{tabular}

\caption{
Effect of candidate pool size. Increasing the beam size enlarges the candidate
pool available to the DPP-inspired selector. Larger candidate pools generally
improve factuality-oriented metrics, particularly FactCC, MiniCheck, and
AlignScore, while reducing ROUGE-Lsum. Best values within each model block
are shown in bold.
}
\label{tab:pool_size}
\end{table*}

Table~\ref{tab:pool_size} examines the effect of candidate pool size by
varying the beam size used during candidate generation. Beam size $4$ serves
as the smallest-pool setting, while beam sizes $8$ and $12$ expose larger
sets of sentence candidates to the DPP-inspired selector.

Larger candidate pools generally improve factuality-oriented performance. On
CNN/DailyMail, increasing the beam size from $4$ to $12$ raises BART
MiniCheck from $94.96$ to $97.73$ and AlignScore from $91.55$ to $94.63$.
For Llama-3-8B, MiniCheck increases from $76.00$ to $87.56$, while
AlignScore increases from $75.08$ to $79.56$. The corresponding FactCC
scores also improve from $75.91$ to $79.93$ for BART and from $46.78$ to
$51.57$ for Llama-3-8B. However, the Llama-based system does not improve
monotonically on FactKB or FaithLens, whose highest values occur in the
beam-$4$ setting.

The benefits are more pronounced on Multi-News. For PRIMERA, increasing the
beam size from $4$ to $12$ improves MiniCheck from $62.26$ to $85.38$,
AlignScore from $52.24$ to $71.27$, and FaithLens from $64.46$ to $79.70$.
FactKB peaks at beam size $8$, indicating that not every factuality metric
improves monotonically with pool size. For Llama-3-8B, all five factuality
metrics improve substantially: MiniCheck rises from $60.46$ to $83.15$,
AlignScore from $59.04$ to $72.56$, FactKB from $90.99$ to $98.29$, and
FaithLens from $68.52$ to $95.14$.

These results indicate that larger candidate pools provide the selector with
more source-supported and complementary content. The gains are accompanied by
lower ROUGE-Lsum, particularly for the Llama-based systems, as well as higher
generation and scoring costs. Candidate pool size therefore acts as a
practical control parameter for balancing source grounding, reference overlap,
and inference efficiency.

\section{Human Evaluation Protocol}
\label{sec:human_protocol}

\begin{table}[!t]
\centering
\small
\renewcommand{\arraystretch}{1.08}
\setlength{\tabcolsep}{3.5pt}
\resizebox{\columnwidth}{!}{%
\begin{tabular}{ll}
\toprule
\textbf{Design Component} & \textbf{Description} \\
\midrule
Evaluation type & Blind pairwise comparison \\
Dataset & CNN/DailyMail \\
Compared systems & Direct generation vs.\ DPP selection \\
Backbones & BART and Llama-3-8B \\
Comparison pairs & 100 total; 50 per backbone \\
Annotators & 20 third-year undergraduate students \\
Pairs per annotator & 5 unique pairs \\
Assignment & One annotator per comparison pair \\
Displayed content & Source, reference, and anonymized outputs \\
Preference labels & A better, B better, Same, Not sure \\
Quality ratings & Five dimensions on a 1--5 scale \\
\bottomrule
\end{tabular}%
}
\caption{Protocol for the blind pairwise human evaluation.}
\label{tab:human_eval_protocol}
\end{table}

We conduct a blind pairwise evaluation on 100 CNN/DailyMail examples:
50 generated with BART and 50 with Llama-3-8B. Twenty third-year undergraduate
annotators each evaluate five unique examples. Each comparison is assigned to
one annotator.

For every example, the annotator is shown the source article, reference
summary, and two anonymized outputs presented in randomized order as
Summary~A and Summary~B. One output is produced by direct generation and the
other by DPP-based selection. System identities are not shown.

Annotators first select one of four overall preference labels:
\emph{A better}, \emph{B better}, \emph{Same}, or \emph{Not sure}. They may
also rate both summaries on a five-point scale according to the following
dimensions:

\begin{itemize}
    \item \textbf{Consistency:} whether the summary's claims are supported by
    the source article;
    \item \textbf{Relevance:} whether the summary captures the most important
    source information;
    \item \textbf{Coherence:} whether the sentences form a logically organized
    and connected summary;
    \item \textbf{Clarity:} whether the summary is readable and easy to
    understand;
    \item \textbf{Conciseness:} whether the summary avoids unnecessary or
    repetitive information.
\end{itemize}

After annotation, the randomized A/B labels are mapped back to the
corresponding systems. Because each comparison is evaluated by one annotator,
inter-annotator agreement cannot be estimated. The results should therefore be
interpreted as a targeted preference study rather than a reliability analysis.

\section{Statistical Significance}
\label{sec:statistical_significance}

We evaluate each beam-12 \texttt{+DPP} system against its same-backbone
direct-generation baseline using paired document-level scores.
For every metric, we report the mean paired difference
$\Delta=\texttt{+DPP}-\text{baseline}$, a 95\% paired-bootstrap confidence
interval based on $10{,}000$ resamples, and a Holm-adjusted $p$ value from a
two-sided paired sign-flip permutation test with $10{,}000$ permutations.
Holm correction is applied over the ten reported metrics within each system
comparison.

\begin{table*}[t]
\centering
\small
\renewcommand{\arraystretch}{1.02}
\setlength{\tabcolsep}{6pt}

\begin{tabular}{@{}lll r c c c@{}}
\toprule
\textbf{Dataset}
& \textbf{Backbone}
& \textbf{Metric}
& $\boldsymbol{\Delta}$
& \textbf{95\% CI}
& $\boldsymbol{p}_{\mathrm{Holm}}$
& \textbf{Sig.}
\\
\midrule

\multirow{20}{*}{\shortstack[l]{CNN/DM\\$n=11{,}490$}}
& \multirow{10}{*}{BART}
& R-1        & $-4.91$  & $[-5.07,\,-4.74]$   & $<10^{-3}$ & \ddag \\
&
& R-2        & $-3.37$  & $[-3.53,\,-3.21]$   & $<10^{-3}$ & \ddag \\
&
& R-L        & $-4.45$  & $[-4.62,\,-4.28]$   & $<10^{-3}$ & \ddag \\
&
& R-Lsum     & $-4.74$  & $[-4.91,\,-4.58]$   & $<10^{-3}$ & \ddag \\
&
& BERTScore  & $-0.99$  & $[-1.02,\,-0.96]$   & $<10^{-3}$ & \ddag \\
\cmidrule(lr){3-7}
&
& FactCC     & $+4.02$  & $[+3.58,\,+4.44]$   & $<10^{-3}$ & \ddag \\
&
& MiniCheck  & $+2.77$  & $[+2.62,\,+2.93]$   & $<10^{-3}$ & \ddag \\
&
& AlignScore & $+3.07$  & $[+2.87,\,+3.28]$   & $<10^{-3}$ & \ddag \\
&
& FactKB     & $+0.99$  & $[+0.79,\,+1.20]$   & $<10^{-3}$ & \ddag \\
&
& FaithLens  & $+1.60$  & $[+1.36,\,+1.85]$   & $<10^{-3}$ & \ddag \\

\cmidrule(lr){2-7}

& \multirow{10}{*}{Llama-3-8B}
& R-1        & $-12.58$ & $[-12.76,\,-12.41]$ & $0.0011$ & \ddag \\
&
& R-2        & $-9.67$  & $[-9.81,\,-9.54]$   & $0.0011$ & \ddag \\
&
& R-L        & $-7.82$  & $[-7.94,\,-7.69]$   & $0.0011$ & \ddag \\
&
& R-Lsum     & $-12.37$ & $[-12.54,\,-12.21]$ & $0.0011$ & \ddag \\
&
& BERTScore  & $-2.66$  & $[-2.70,\,-2.63]$   & $0.0011$ & \ddag \\
\cmidrule(lr){3-7}
&
& FactCC     & $+4.81$  & $[+4.16,\,+5.44]$   & $0.0011$ & \ddag \\
&
& MiniCheck  & $+11.56$ & $[+11.18,\,+11.95]$ & $0.0011$ & \ddag \\
&
& AlignScore & $+4.48$  & $[+4.08,\,+4.88]$   & $0.0011$ & \ddag \\
&
& FactKB     & $+0.04$  & $[-0.18,\,+0.24]$   & $0.745$  & n.s. \\
&
& FaithLens  & $-0.21$  & $[-0.46,\,+0.04]$   & $0.248$  & n.s. \\

\midrule

\multirow{20}{*}{\shortstack[l]{Multi-News\\$n=5{,}622$}}
& \multirow{10}{*}{PRIMERA}
& R-1        & $-4.41$  & $[-4.65,\,-4.18]$   & $<10^{-3}$ & \ddag \\
&
& R-2        & $-4.01$  & $[-4.19,\,-3.83]$   & $<10^{-3}$ & \ddag \\
&
& R-L        & $-2.61$  & $[-2.78,\,-2.45]$   & $<10^{-3}$ & \ddag \\
&
& R-Lsum     & $-4.50$  & $[-4.72,\,-4.27]$   & $<10^{-3}$ & \ddag \\
&
& BERTScore  & $-2.41$  & $[-2.46,\,-2.36]$   & $<10^{-3}$ & \ddag \\
\cmidrule(lr){3-7}
&
& FactCC     & $+8.28$  & $[+7.55,\,+9.03]$   & $<10^{-3}$ & \ddag \\
&
& MiniCheck  & $+22.32$ & $[+21.76,\,+22.88]$ & $<10^{-3}$ & \ddag \\
&
& AlignScore & $+17.97$ & $[+17.43,\,+18.53]$ & $<10^{-3}$ & \ddag \\
&
& FactKB     & $+3.15$  & $[+2.71,\,+3.58]$   & $<10^{-3}$ & \ddag \\
&
& FaithLens  & $+15.24$ & $[+13.80,\,+16.70]$ & $<10^{-3}$ & \ddag \\

\cmidrule(lr){2-7}

& \multirow{10}{*}{Llama-3-8B}
& R-1        & $-6.60$  & $[-6.79,\,-6.40]$   & $<10^{-3}$ & \ddag \\
&
& R-2        & $-3.26$  & $[-3.35,\,-3.17]$   & $<10^{-3}$ & \ddag \\
&
& R-L        & $-2.23$  & $[-2.32,\,-2.14]$   & $<10^{-3}$ & \ddag \\
&
& R-Lsum     & $-6.62$  & $[-6.80,\,-6.45]$   & $<10^{-3}$ & \ddag \\
&
& BERTScore  & $-1.82$  & $[-1.86,\,-1.78]$   & $<10^{-3}$ & \ddag \\
\cmidrule(lr){3-7}
&
& FactCC     & $+15.35$ & $[+14.55,\,+16.15]$ & $<10^{-3}$ & \ddag \\
&
& MiniCheck  & $+22.67$ & $[+22.12,\,+23.24]$ & $<10^{-3}$ & \ddag \\
&
& AlignScore & $+13.52$ & $[+12.99,\,+14.05]$ & $<10^{-3}$ & \ddag \\
&
& FactKB     & $+7.30$  & $[+6.74,\,+7.87]$   & $<10^{-3}$ & \ddag \\
&
& FaithLens  & $+26.63$ & $[+25.31,\,+27.91]$ & $<10^{-3}$ & \ddag \\

\bottomrule
\end{tabular}

\caption{
Paired statistical comparisons of the four beam-12 \texttt{+DPP} systems
against their same-backbone direct-generation baselines in
Table~\ref{tab:main_results}. $\Delta$ is computed as
\texttt{+DPP} minus baseline and is reported in percentage points.
Confidence intervals are obtained from $10{,}000$ paired-bootstrap resamples.
Holm-adjusted $p$ values are computed using two-sided paired sign-flip
permutation tests with $10{,}000$ permutations, with correction over the ten
metrics within each system comparison. \ddag{} denotes
$p_{\mathrm{Holm}}<0.01$, and n.s.\ denotes a nonsignificant difference.
Because $\Delta$ is computed from paired document-level scores, it may differ
slightly from differences between the rounded aggregate values in
Table~\ref{tab:main_results}.
}
\label{tab:significance_with_faithlens}
\end{table*}

Table~\ref{tab:significance_with_faithlens} confirms the trade-off observed in
the main results. All four \texttt{+DPP} systems show significant decreases in
ROUGE and BERTScore, together with significant improvements on nearly all
factuality metrics. On CNN/DailyMail, \texttt{BART+DPP} improves FactCC by
$4.02$ points, AlignScore by $3.07$ points, and FaithLens by $1.60$ points.
The gains are substantially larger on Multi-News:
\texttt{PRIMERA+DPP} improves MiniCheck by $22.32$ points and AlignScore by
$17.97$ points, while \texttt{Llama-3-8B+DPP} improves MiniCheck by $22.67$
points and FaithLens by $26.63$ points. The only nonsignificant differences are
FactKB and FaithLens for \texttt{Llama-3-8B+DPP} on CNN/DailyMail. Overall,
the paired analyses support the conclusion that global sentence-level
selection reliably improves automatic source-grounding metrics, particularly
in multi-document summarization.

\end{document}